\documentclass[letterpaper, 10 pt, conference]{ieeeconf}

\IEEEoverridecommandlockouts                              

\usepackage{multicol}

\usepackage{xcolor}

\usepackage{times}
\usepackage{epsfig}
\usepackage{graphicx}
\usepackage{tabularx}
\usepackage{amsmath}
\usepackage{amssymb}
\usepackage{bbm}
\usepackage[misc]{ifsym}
\usepackage[export]{adjustbox}
 \usepackage{multirow}

\usepackage{pgf}

\usepackage{graphics}
\usepackage{pifont}

\usepackage{times}
\usepackage{soul}
\usepackage{url}
\usepackage{graphicx}
\usepackage{amsmath}
\usepackage{booktabs}
\usepackage{balance} 
\usepackage{wrapfig}
\usepackage{fancybox}
\usepackage{tikz}
\usepackage{algpseudocode}
\usepackage{algorithmicx}
\usepackage[ruled,vlined,linesnumbered]{algorithm2e}
\usepackage{diagbox}
\usepackage{amsfonts}       

\usepackage{graphbox}

\graphicspath{{figures/}}

\usepackage[colorlinks=true,bookmarks=true]{hyperref}

\begin{document}
\title{\Large \textbf{ SEAL: Simultaneous Exploration and Localization in Multi-Robot Systems }}
\author{Ehsan Latif \and Ramviyas Parasuraman
\thanks{The authors are with the Heterogeneous Robotics Research Lab, School of Computing, University of Georgia, Athens, GA 30602, USA}
\thanks{Author emails: \tt{\{ehsan.latif;ramviyas\}}@uga.edu}
}

\maketitle              

\begin{abstract}
The availability of accurate localization is critical for multi-robot exploration strategies; noisy or inconsistent localization causes failure in meeting exploration objectives. We aim to achieve high localization accuracy with contemporary exploration map belief and vice versa without needing global localization information. This paper proposes a novel simultaneous exploration and localization (SEAL) approach, which uses Gaussian Processes (GP)-based information fusion for maximum exploration while performing communication graph optimization for relative localization. Both these cross-dependent objectives were integrated through the Rao-Blackwellization technique. Distributed linearized convex hull optimization is used to select the next-best unexplored region for distributed exploration. SEAL outperformed cutting-edge methods on exploration and localization performance in extensive ROS-Gazebo simulations, illustrating the practicality of the approach in real-world applications.
\end{abstract}

\section{Introduction}
\label{sec:intro}
The growing use of autonomous robots in various applications, including environmental monitoring, search and rescue, and mapping, has made multi-robot robotic exploration a major study field in recent years. In these scenarios, the robot must explore uncharted territory while simultaneously maintaining other criteria such as map accuracy, cost of travel, travel time, and energy savings. Information-based exploration methods have drawn particular attention because of their capacity to facilitate faster exploration and efficiently expand to 3D scenes. Many studies have examined various methods to achieve these objectives. To build a reward function and choose the best control strategies that reduce the uncertainty of maps, these techniques employ information-theoretic metrics like mutual information (MI) \cite{xu2021crmi}.

\begin{figure}[t]
\centering
 \includegraphics[width=0.9\linewidth]{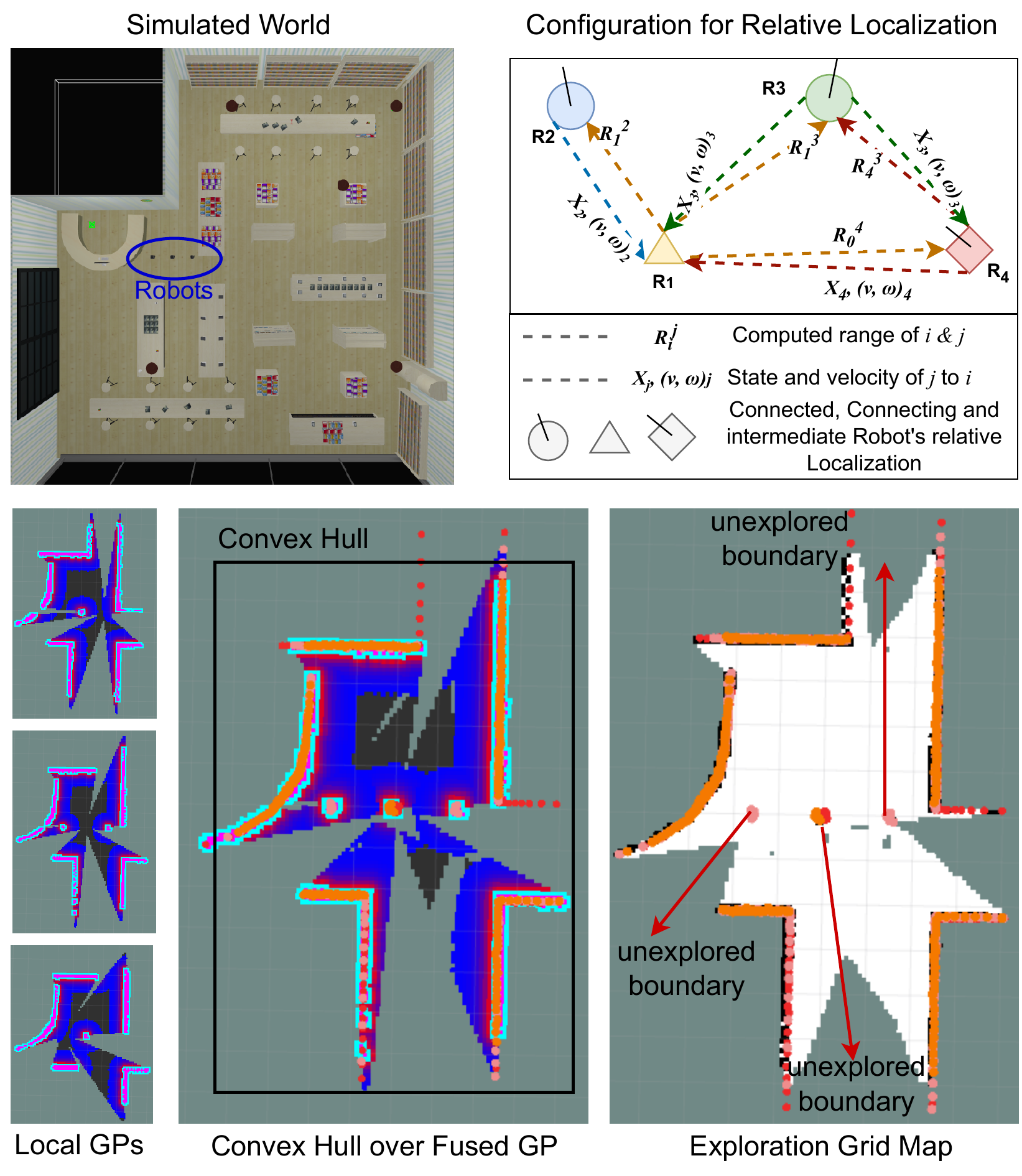}
 \caption{Overview of the proposed SEAL for simultaneous exploration and localization in the multi-robot system, shown with the ROS Gazebo simulation, configuration for relative localization same as DGORL \cite{latif2022dgorl}, and the exploration process with the convex hull}
 \label{fig:overview}
\end{figure}

Entropy \cite{botteghi2020entropy}, mutual information \cite{jadidi2015mutual}, and Bayesian optimization \cite{bai2016information} based multi-robotic exploration techniques are only a few of the many information-based exploration techniques proposed in the literature. To reduce map uncertainty when exploring new areas, these techniques improve the exploration process based on several information-theoretic measures. Specifically, reducing pose uncertainty can significantly reduce map uncertainty, and an information-based controller that takes pose uncertainty reduction into account can direct the robot to a location where it is more likely to find a loop closure, significantly improving the overall accuracy of the trajectory and the resulting map \cite{valencia2018active}.

Accurate positional data is required for multi-robot exploration, which is one of the challenges posed in GPS-denied environments. Conventionally, the robot employs sensors to measure the surroundings and updates its position according to the sensors' data. This strategy can fail if the sensor is faulty or does not provide enough information. 
In these circumstances, simultaneous localization and mapping (SLAM) methods can be applied \cite{schmuck2019ccm}, where the robot estimates its position in the environment and simultaneously creates a map of the surrounding area.
While the existing multi-robot SLAM approaches in the literature can handle dynamic scenes very well, they might not work well in contexts with few map features or without prior global position information \cite{latif2022multi}. Furthermore, SLAM requires feature-based map merging, which is a computationally expensive procedure and causes failure during the expansion of the global map. However, we aim to overcome this limitation by only applying efficient GP fusion in parallel to the relative localization, which makes the overall system computationally efficient and provides accurate positioning and mapping information.

This paper proposes a SEAL approach incorporating information-based exploration techniques to enhance the efficiency and accuracy of robotic localization and exploration designed for multi-robot systems.  To choose the next-best viewpoint for exploration, we use linearized convex hull optimization, which ensures maximum coverage of the unknown region. Our proposed method uses a Gaussian process (GP) modeling based on the fusion of multiple GPs received from connected robots and relative position estimation, further improving the posterior belief of exploration and localization. Furthermore, our method is appropriate for dynamic settings with limited features or when there is no prior knowledge and does not rely on external sensors. Fig.~\ref{fig:overview} provides an overview of the proposed method implemented in the Robot Operating System (ROS) framework.

The main contributions of SEAL are as follows:
\begin{itemize}
    \item We propose a new distributed exploration strategy for multi-robot systems using Gaussian process (GP) modeling over relative localization with inter-robot communication data. Here, we integrate the Rao-Blackwellization technique to improve the posterior beliefs of both exploration and localization.
    \item We focus on maximizing the coverage (exploration) of an unknown environment with multiple robots using the linearized convex hull optimization method.
    \item We extensively evaluate our SEAL approach's exploration performance in realistic ROS Gazebo simulations compared against two state-of-the-art exploration approaches: Rapidly exploring Random Tree (RRT)-based exploration \cite{zhang2020rapidly} and Deep Reinforcement Learning (DRL)-based exploration \cite{hu2020voronoi}.
    \item To further analyze the performance of localization, we compare with typical Gmapping-based SLAM (used in RRT Exploration \cite{zhang2020rapidly}) and a relative localization method, DGORL \cite{latif2022dgorl}, which is based on our prior work.
    \item We open source\footnote{\url{https://github.com/herolab-uga/ROS-SEAL}} our method as a ROS package for use and further development by the robotics community.
\end{itemize}

The SEAL's motivation is utilizing data sharing through inter-robot communication to simultaneously localize and explore the environment. It continually improves the beliefs of exploration and localization using Rao-Blackwellization for environments where global position information is unknown. Furthermore, ensuring maximum exploration inspired by convex hull optimization \cite{pham2009convex} and its linearization addresses the problem of coverage in unknown and constrained environments. 
Our method does not rely on global information for localization and boundary information for exploration.
SEAL achieved high localization, exploration accuracy, and efficiency by employing parallel computation of localization and exploration powered by relative localization \cite{latif2022dgorl}, Rao-Blackwellization \cite{rao2008rao}, and linearized convex hull optimization to ensure maximum area coverage.

\section{Related Work}
\label{sec:relatedwork}
Occupancy grid maps are extensively used in 2D and 3D settings because they can be swiftly queried and updated \cite{hornung2013octomap}. Additionally, there are several applications where the data in OGMs can be used for navigation and collision-free path planning. By utilizing these benefits, OGMs offer a new way to quantify map uncertainty in light of newly observed data \cite{zhang2020fsmi}.

Collaborative simultaneous localization and mapping (CSLAM), which is possible with a centralized system, is an alternative strategy to maximize the effectiveness of exploration. 
Shi et al. \cite{shi2020adaptive} presented an adaptive informative sampling strategy that divides the environment into various sections. This strategy, however, necessitates a substantial amount of communication and computational resources. Placed and Castellanos \cite{placed2021fast} suggested a quick autonomous robotic exploration strategy using an underlying graph structure. Their approach makes it possible to explore new settings quickly and effectively. All of these approaches use OGMs as a source of mapping and exploration. 

Frontier detection algorithms are built for fast navigation using rapidly expanding random trees known as RRTs. For instance, to improve the effectiveness and efficiency of multi-robot map exploration, the aim is integrated into an optimization framework that uses RRTs in \cite{zhang2020rapidly}. However, this strategy's limitations are the cost of computing the optimization algorithms and the potential for non-optimal results due to the stochastic nature of RRTs.

OGMs are not the only mapping method; continuous occupancy mapping is one (COM). For instance, given the observed observations (training set) comprising free and occupied space, Gaussian processes-based occupancy maps \cite{swiler2020survey} employ kernel inference methods to learn distributions across continuous occupancy. GP maps may query map points at any resolution and record the spatial correlations between sample points. They are particularly useful for extrapolating the occupancy and uncertainty of unseen places.
For instance, an adaptive semantic perception strategy for quick and secure outdoor exploration was presented by Wang et al. \cite{wang2022fast}. Their approach enhances a robot's ability to adapt to shifting settings. 
Recently, for robotic exploration, Xu et al. \cite{xu2021crmi} put forth a confidence-rich localization and mapping strategy based on a particle filter. Their method increased the robot's stability and localization accuracy under challenging situations.
%

Recent Bayesian kernel inference-based efforts, including \cite{gan2020bayesian}, offer more computationally effective continuous occupancy mapping techniques.
Jadidi et al. \cite{jadidi2014exploration} presented a GP-based mapping and exploration technique (GPMI) to identify frontiers defined by mean occupancy from the GP map to realize COM-based information-theoretic exploration. In the extended work by Jadidi et al., \cite{ghaffari2018gaussian}, a high-dimensional logistic regression classifier used to produce a probabilistic frontier representation based on the uncertainty implied by GP is also noteworthy. The forward sensor model (FSM) and combined predictive distribution over the entire map are used to numerically design and compute the GPMI surfaces in the present perception field.

Another method that is gaining popularity in solving exploration and mapping difficulties is reinforcement learning (RL). A deep RL technique for autonomous graph exploration with fallible sensing capabilities was presented by Chen et al. \cite{chen2020autonomous}. Their technique performed well but can only be used in relatively narrow surroundings. Chen et al. \cite{chen2021zero} also proposed a zero-shot RL technique on graphs for autonomous exploration in uncertain scenarios. Although their method drastically reduced training time, it hasn't been tested in real-world situations environmental changes may hamper robotic exploration in outdoor settings.
Moreover, a Voronoi-based strategy that uses DRL for cooperative multi-robot exploration is suggested in \cite{hu2020voronoi}. The study aims to improve exploration efficiency by dividing the environment into Voronoi cells and assigning a robot to each cell for study. However, this approach is constrained by the difficulty of training deep RL models and the potential for less-than-ideal results because of the complexity of the environment and the unknown mapping of the environment.

The proposed SEAL approach is compared to frontier detection-based RRT \cite{zhang2020rapidly} and DRL-based \cite{hu2020voronoi} exploration strategies. Both compared approaches are assumed to have accurate position information to be fully functional; robots merge locally examined maps and share them between iterations, increasing computing complexity. 
SEAL overcomes limitations of existing solutions of explorations that require accurate global positioning information and works on feature matching map merging techniques by providing an optimal solution. 

\section{Problem Formulation}
\label{sec:problem}
\textbf{Problem Statement:} Given that $n$ robots operating in an uncharted region are wirelessly connected, using shared exploration grids and relative location information, robots should efficiently and accurately perform simultaneous exploration and localization.

\textbf{Notations:} Let $r_i$ be the concerning robot, $n$ be the total number of robots deployed in the unknown environment, and $X_i$ be the robot's position. We only consider $x,y$ coordinates of robots from the frame of reference of $r_i$, which is sufficient to update the exploration grid. Let $z_{i,j}$ be the scanned observation between robots $i$ and $j$ and $g_i$ be the exploration grid for robot $i$. Additionally, $w_i$ is the robot's weight at $i^{th}$ exploration grid $g_i$, $s^i_j$ is the observed Wi-Fi Radio Signal Strength Indicator (RSSI) for connected robot $j$, and $C_i$ is the convex hull of the robot's postures.

The difficulty lies in figuring out the robots' positions while minimizing the sensing gap between them and building an accurate grid map of the bounding box. The problem is best described as follows:
\begin{equation}
    \min_{X_{1:n}} \sum_{i=1}^{n} \sum_{j \in \mathcal{N}_i} \text{dist}(X_i,X_j)
\end{equation}
Each robot $r_i$ forms a 2-D Pose Estimation Graph (PEG) $G_i=(V_i, E_i)$ using the Received Signal Strength Indicator (RSSI) from connected robots. $\Tilde{X_i}$ is the position estimation of the objective function:
\begin{equation}
    \min_{\Tilde{X_{1:n}}} \sum_{i=1}^{n} \sum_{j \in \mathcal{N}_i} (\Tilde{X_i} - X_j - z_{i,j})^2
\end{equation}
Robot $r_i$ forms an exploration grid considering uniform distributions over the workspace and receives observations $z_{i,j}$ from the connected robots at each position and updates its global grid as $ g_{global} = \sum_{i=1}^{n} w_i g_i$. Further, $\Tilde{X_i}$ is updated using Rao-Blackwellization \cite{robert2021rao} for the updated belief state:
\begin{equation}
    p(\Tilde{X_i},g_i|z_{i,j}) = p(\Tilde{X_i}|z_{i,j})p(g_i|\Tilde{X_i})
\end{equation}
Lastly, the convex hull $C_i$ is updated using linearized convex hull optimization for the following objective :
\begin{equation}
    \min_{x_{1:n}} \sum_{i=1}^{n} \sum_{j \in \mathcal{N}_i} \text{dist}(\Tilde{X_i},\Tilde{X_j})
\end{equation}

As the above convex hull objective function is non-linear, it can be linearized as:
\begin{equation}
    \begin{aligned}
\text{minimize} &\quad f(\Tilde{X_i}) \ \text{subject to} &\quad \Tilde{X_i} \in X, 
\end{aligned}
\end{equation}

Where $f(\Tilde{X_i})$ is the global objective function, $\Tilde{X_i}$ is the robots' positions in the region of low wireless vicinity, and $X$ is the set of feasible positions.
\begin{equation}
    f(\Tilde{X_i}) \approx f(\Tilde{X_i}) + \nabla f(\Tilde{X_i})^T(\Tilde{X}-\Tilde{X_i})
\end{equation}

 $\nabla f(\Tilde{X_i})$ is the gradient of the objective function evaluated at $\Tilde{X_i}$.

\begin{figure*}[t]
    \centering
\includegraphics[width=0.98\linewidth]{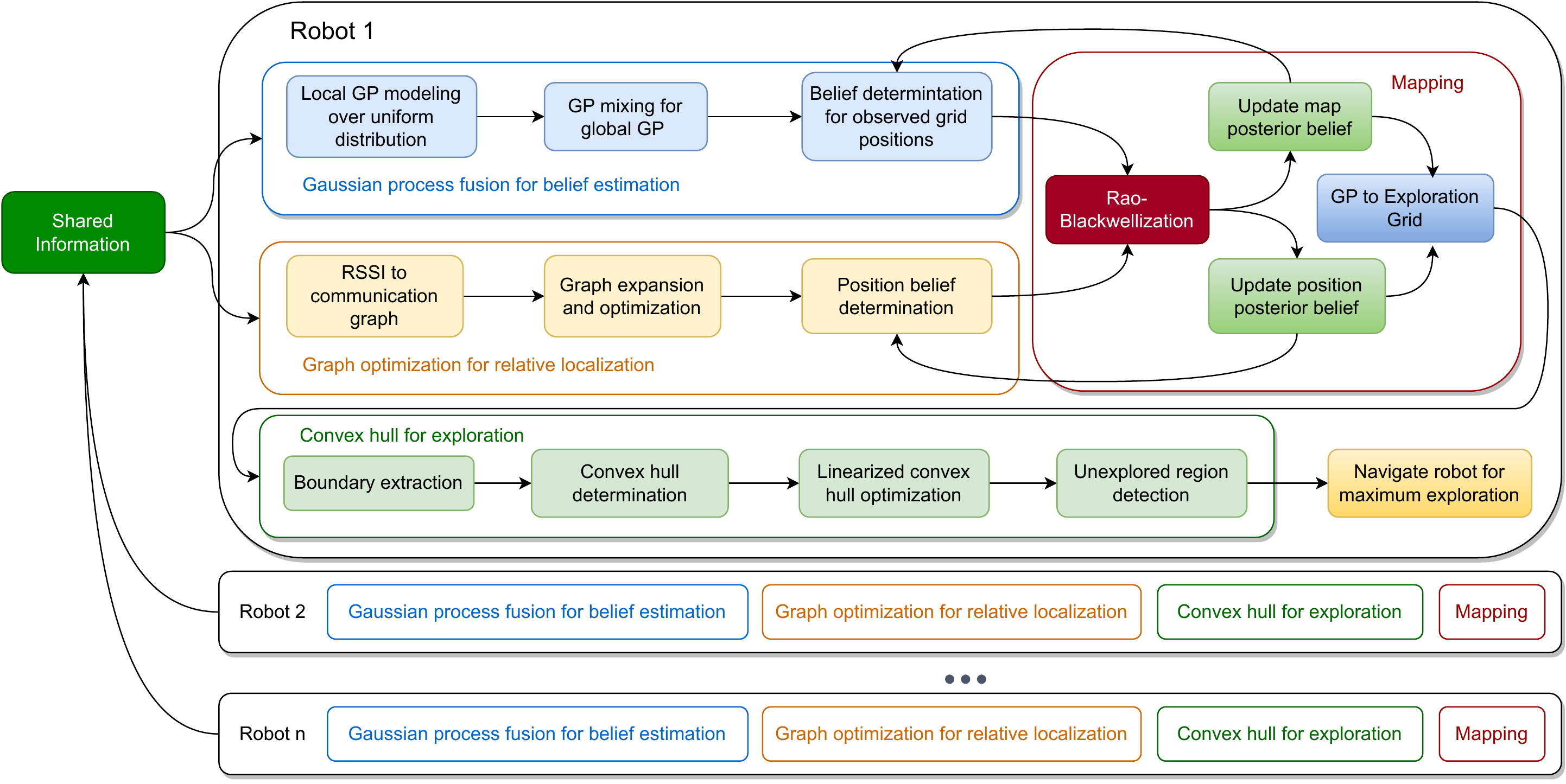}
\caption{System architecture of SEAL distributed across several robots. It shows Robot 1's process of exploring, convex hull optimization, and relative localization operations using shared information from $n$ connected robots.}
    \label{fig:system-architecture}
    \vspace{-2mm}
\end{figure*}

To maximize the coverage, the linearized objective function can be minimized to estimate the following position to navigate for $r_i$. Robots then position themselves to cover the closed region in the bounded area efficiently. Standard barrier certificates and navigation stacks are utilized for safe autonomous navigation. The system architecture of SEAL for the mentioned problem can be visualized in Fig.~\ref{fig:system-architecture}.

\begin{algorithm}[tb]
\SetAlgoLined
\caption{Distributed SEAL}
\label{alg:SEAL}
\textbf{Required:} previous position $X_{t-1}$, local GP $g$, received GPs, RSSI from connected robots $S$, number of connected robots $n$\;
\SetKwBlock{DoParallel}{do in parallel}{end}
    \DoParallel{
    $g$ = Local predicted GP grid\;
    \For{$g_j \in GPs$ from connected robots}{
    Merge received $g_j$ into local $g$\;
    $b_j^g$ = update exploration belief of $g$ as Eq.~\ref{eqn:gp_belief}\;
    }
    \For{$j \in$ connected robots}{
    Apply relative localization using DGORL \cite{latif2022dgorl}\;
        $bel_j(\Tilde{X})$ =  update position belief\;
    }
    
    }
Apply Rao-Blackwellization to update exploration as well as position belief\;
Update position belief using GP belief as Eq.~\ref{eqn:update_pos_belief}\;
Update GP belief by incorporating Entropy $H(bel(\title{X}))$ from Eq.~\ref{eqn:entropy}\;
Build occupancy grid map based on updated belief from GP\;
$C$ = received linearized convex hull using Alg.~\ref{alg:linear_convex_hull}\;
Navigate the robot to the unexplored region based on $C$ for max exploration.
\end{algorithm}

\section{Approach and Algorithms}
\label{approach}
SEAL is composed of four major parts:
\begin{enumerate}
    \item Global Exploration over Distributed Gaussian Process (GP) Modeling \cite{luo2018adaptive}.
    \item Probabilistic Extension of Relative localization using DGORL \cite{latif2022dgorl}.
    \item Integration of relative localization with exploration to update belief state using modified Rao-Blackwellization.
    \item Multi-robot distribution for full coverage using linearized convex hull optimization.
\end{enumerate}
The method proposed in this work leverages the strengths of non-parametric generalization of GP for individual robot exploration, polynomial time graph optimization for relative localization, probabilistic efficiency of Rao-Blackwellization to update belief state, and linearized convex hull optimization for complete coverage without performing Voronoi partitioning. Our proposed SEAL is a merger of relative location with exploration, along with the linearized extension of the convex hull algorithm that can leverage multi-robot teams to cover the bounded region efficiently and comprehensively.

Furthermore, SEAL relies only on shared data between robots for position estimation and exploration, without considering any centralized information sources. Alg.~\ref{alg:SEAL} elaborates the distributed approach of SEA utilizing GPs and relative localization belief information to generate a map and determine the unexplored region for the robot to navigate to using linearized convex hull optimization.

\subsection{Gaussian Process for Exploration}
\label{gaussian_process}
GP regression is a method that is frequently used to model spatial phenomena. Simulating the hidden mapping from training data to the target phenomenon is possible while considering the uncertainty provided by the natural non-parametric generalization of linear regression. Assume that the target phenomenon, a wireless source, satisfies a multivariate joint Gaussian distribution in this case. The Gaussian probability distribution of the phenomena $\omega(g)$ as given by the mean function $\mu(g)=\varepsilon[(g)]$ and covariance function $\delta^2(g,g')=\varepsilon[(\omega(g)-\mu(g))^T\times(\omega(g)-\mu(g)]$ is the output of the trained GP model using training data.

Formally, let $GP^{[i]}=[g^{[i]}_1,...,g^{[i]}_{k}]^T$ be the grid positions at which $r_i$ observed $k$ noisy RSSI readings \cite{parasuraman2014multi} from the source $S^{[i]}=[s^{[i]}_1,...,s^{[i]}_{k}]^T$. Since the mean function is assumed to be zero without losing generality, each observation is noisy, with the formula $y=\omega(g)+ \epsilon$. To achieve this, given a sampling location $g_{k}\in GP$, we have the conditional posterior means $\mu_{g_{k}|GP_j,y_j}$ and variance $\delta^2_{g_{k}|GP_j,y_j}$ as follows from the learned GP model defining the Gaussian distribution of $\omega(g_{k})\sim N(\mu_{g_{k}|GP_j,y_j},\delta^2_{g_{k}|GP_j,y_j})$ for sample position $k$.

Once each robot learns its GP model for its current position, they share GPs with the connected robots for global exploration. Since each local GP is assumed to be learned by each robot, we assume that the underlying data distribution to be explored can be characterized by a combination of $n$ fused GPs, $GP=\{GP_1,..., GP_n\}$ received from connected robots. We define $p(g \mid s_{1:n},X_{1:n}) = b_n^g$ as the likelihood that, for each point in the environment, $g \in GP$ is best characterized by the GP that the $r_i$ has learned. Then, we produce a combination of GP models weighted by the $b_n^g$ at any grid $g\in GP$ as a linear combination of ${GP_1,..., GP_n}$. The means $\mu'_{g|GP,Y}$ and variance $\delta'^2_{g|GP,Y}$ for robot $i$ are the parameters that define the model as:
\begin{equation}
\begin{aligned}
& \mu'^{[i]}_{g\mid GP,Y} = \sum\limits_{j = 1}^n b_j^{g^{[i]}}\times{\mu^{[i]}_{g\mid {g_j},{{y}_j}}}\\
& \delta'^{2^{[i]}} _{g\mid GP,Y} = \sum\limits_{j = 1}^n b_j^{g^{[i]}}\times \left( {\delta _{g\mid {g_j},{{y}_j}}^{2^{[i]}} + {{\left( {{\mu^{[i]}_{g\mid {g_j},{y_j}}} - \mu'^{[i]}_{g\mid GP,Y}} \right)}^2}} \right)
\end{aligned}
\end{equation}
In this problem, the algorithm computes the weight distribution in the expectation step (E-Step). Then, the maximization step updates the parameters of the local fused GP with the estimated weight distribution (M-Step), the same as discussed in GPMix \cite{shi2020adaptive}. Finally, the weight distribution is set to the following values before the first iteration for each random wireless sample $s$ at grid position $g$:
\begin{equation}
O_{b^{g}}
\approx 
\left\{ {\begin{array}{ll} 1&{{\text{ if }}{g}\in {GP^{[i]}}}, b^{g}\geq \theta \\
0 & {{\text{ Otherwise }}} \end{array}\quad \forall j = 1, \ldots ,n} \right.\
\label{eqn:gp_belief}
\end{equation}
Where $\theta$ is the threshold for exploration and $O_{b^{g}}$ is the value of grid position in the exploration grid map. 


\subsection{Relative Localization}
\label{relative_localization}
Assume at a given time $t$, a team of robots considered as vertices of the graph contains $n \in \mathbb{N}$ connected robots can form a weighted undirected graph, denoted by $G=(V, E, A)$, of order $n$ consists of a vertex set $V=\{v_1,...,v_n\}$, an undirected edge set $E \in V \times V$ is a range between connected robots and an adjacency matrix $A=\{a_{ij}\}_{n \times n}$ with non-negative element $a_{ij}>0$, if $(v_i,v_j) \in E$ and $a_{ij}=0$ otherwise. 
An undirected edge $e_{ij}$ in the graph $G$ is denoted by the unordered pair of robots $(v_i,v_j)$, which means that robots $v_i$ and $v_j$ can exchange information with each other. 

Here, we only consider the undirected graphs, indicating that the robots' communications are all bidirectional. Then, the connection weight between robots $v_i$ and $v_j$ in graph $G$ satisfies $a_{ij}=a_{ji}>0$ if they are connected; otherwise, $a_{ij}=a_{ji}=0$. Without loss of generality, it is noted that $a_{ii}=0$ indicates no self-connection in the graph. The degree of robot $v_i$ is defined by $d(v_i) = \sum_{j=1}^{n} a_{ij}$ where $j \neq i$ and $i=1,2,...,n$. The Laplacian matrix of the graph $G$ is defined as $L_n=D-A$, where $D$ is the diagonal with $D=diag\{d(v_1),d(v_2),...,d(v_n)\}$. If the graph $G$ is undirected, $L_n$ is symmetric and positive semi-definite. A path between robots $v_i$ and $v_j$ in a graph $G$ is a sequence of edges $\{(v_i,v_{i1}),(v_{i1},v_{i2}),...,(v_{in-1},v_{in}),(v_{in},v_j)\}$ in the graph with distinct robots $v_{in} \in V$. An undirected graph $G$ is connected if a path exists between any pair of distinct robots $v_i$ and $v_j$ where $(i,j=1,...,n)$.

We begin by creating the Estimated Relative Position Measurement Graph (ERPMG):  $G_E = (V_E, E_E, A_E )$, based on the Relative Position Estimation Graph (RPMG): $G = (V, E, A)$, which can be constructed using range information received from connected robots of an n-robot system and describes the relative position measurements among robots. Based on robotic motion constraints, we extend RPMG to accommodate all possible robot positions in the succeeding time step.

Since each robot also gets RSSI from other robots, we may map the predicted relative positions for each robot using the model and identify the $k$ soft maximum out of them by locating the intersection region as an area of interest. Once we have $k$ possible positions of each robot, we can generate $<n^k$ solvable graphs for optimization.

We consider a network with $n$ robots for optimization of expanded graphs, labeled by $V=\{1,2,...,n\}$ and $k$ possible connections to other robots. Every robot $i$ has a local convex objective function and a global constraint set. The network cost function is given by:
\begin{align*} 
\mathrm {minimize} ~~&f( \mathbf {x}) = \sum \limits _{i=1}^{N} f_{i}( \mathbf {x}) \\ 
\mathrm {subject~to} ~~& \mathbf {x}\in \mathcal {D}= \left \{{ \mathbf {x}\in \mathbf {R}^{k}~:~c( \mathbf {x}) \leq 0 }\right \}
\label{eqn: constraints}
\end{align*}
Here, $\mathbf{x}\in R^k$ is a global decision vector; $f_i: R^k\to R$ is the convex objective function of robot $i$ known only by robot $i$; $D$ is a bounded convex domain, which is (without loss of generality) characterized by an inequality constraint, i.e., $c(\mathbf{x})\leq 0$, where $c: R^k\to R$ is a convex constraint function known by all robots.
In addition, to estimate the relative positioning of connected robots, we also calculate the probability of estimation $bel_j(\Tilde{X}) = p(\Tilde{X}_j|\Tilde{X}_i,s_{i,j})$ for each relative position, where $\Tilde{X}$ corresponds to the position estimation by $\mathbf{x}$, which will be used in the improvisation of the exploration grid. Technical and theoretical details of the proposed relative localization technique are explained in our previous work DGORL \cite{latif2022dgorl}.

\subsection{Rao-Blackwellization for SEAL}
\label{sec:rao-blackwell}
In the Rao-Blackwellization update procedure, according to \cite{rao2008rao}, the proposal distribution is suboptimal, especially when the proprioceptive odometer measurements are less accurate than the position estimates through the relative location as mentioned in Section \ref{relative_localization}. Instead,  we can use the persistent exploration grid map belief in Gaussian Process as discussed in Section \ref{gaussian_process}  to improve the localization accuracy and vice versa. Furthermore, in a  Bayesian  framework,  we  explicitly  incorporate  the exploration grid map belief into the measurement likelihood function:
\begin{equation}
\begin{aligned} 
& bel_{j}^{[i]}(\Tilde{X}) \\ 
& =\psi \overline{bel}_{j}^{[i]}(\Tilde{X}) \int_{g} p(s_{j} \vert g, \Tilde{X}_{1: j}, s_{1: j-1})^{[i]} p(g \vert \Tilde{X}_{1: j-1}, s_{1: j-1})^{[i]} dg \\ 
& =\psi \overline{bel}_{j}^{[i]}(\Tilde{X}) \int_{g} p(s_{j} \vert g, \Tilde{X}_{1: j}, s_{1: j-1})^{[i]} b_{j-1}^{{g}^{[i]}} dg
\end{aligned}
\label{equ:belief}
\end{equation}

According to the definition of posterior map belief, $b_{j-1}^{g^{[i]}}$ is a sufficient statistic for all previous positions $\Tilde{X}_{1:j-1}$ and RSSI $s_{1:j-1}$, and $\psi$ is the belief update constant, the following expression holds:
\begin{equation} 
p(s_{j} \vert g, \Tilde{X}_{1: j}, s_{1: j-1})^{[i]} \approx p(s_{j} \vert g, \Tilde{X}_{j}, b_{j-1}^{g})^{[i]}
\end{equation}
Similarly, the belief $b_{j}^{g^{[i]}}$ is also a sufficient statistic for the current estimated position $\Tilde{X}_j$ and the previous map belief $b_{j-1}^{g^{[i]}}$ , so we can get:
\begin{equation}
p(s_{j} \vert g, \Tilde{X}_{j}, b_{j-1}^{g})^{[i]} \approx p(s_{j} \vert b_{j}^{g})^{[i]}
\end{equation}
Thus Eq.~\ref{equ:belief} can be written as:
\begin{equation}
bel_{j}^{[i]}(\Tilde{X})\propto\overline{bel}_{j}^{[i]}(\Tilde{X})\int_{g}p(s_{j}\vert b_{j}^{g})^{[i]} b_{j-1}^{g^{[i]}}dg
\label{eqn:update_pos_belief}
\end{equation}
Now the weight of sample $j$ possesses the sample position for robot $i$ can be defined as:
\begin{equation}
w_{j}^{[i]} = \int_{g}p(s_{j}\vert b_{j}^{g^{[i]}})b_{j-1}^{g^{[i]}}dg. 
\end{equation}
Furthermore, for a sample approximating the position belief, a straightforward method to estimate pose uncertainty is to use all normalized weights in a discretized way as:
\begin{equation}
\begin{aligned} 
H({bel}^{[i]}(\Tilde{X})) & = -\int bel^{[i]}(\Tilde{X}) \log bel^{[i]}(\Tilde{X}) d \Tilde{X}\\
&\approx-\sum_{j=1}^{N_{s}} \omega_{j}^{[i]} \log \omega_{j}^{[i]}
\end{aligned}
\label{eqn:entropy}
\end{equation}

\begin{algorithm}[tb]
\SetAlgoLined
\caption{Distributed Linearized Convex Hull Optimization}
\label{alg:linear_convex_hull}
C = convex hull $(O_f \cup O_o)$ \texttt{\#convex hull of observed walls and space}\;
$C_o = O_o \in H$ \texttt{\#identify occupied cells on convex hull contour}\;
$L_o$ = Probabilistic Hough Lines$(C_o)$ \texttt{\#search for lines on convex hull}\;
\For{each $l$ in $L_o$} {
    $c_{i,j} \in I_o$ = find intersections$(l)$ \texttt{\#identify intersecting hull lines}\;
}
C = convex hull$(O_f \cup O_o \cup I_o)$  \texttt{\#convex hull of observations and intersection points}\;
\eIf{C is linear}{
    \Return C\;
}{
    $L_C$ = Linearize (C)  \texttt{\#Linearize convex hull}\;
    \Return $L_C$\;
}
\end{algorithm}

\subsection{Maximum Exploration with Distributed Linearized Convex Hull Optimization}
The primary goal of linearizing convex hull optimization is to maximize the coverage of the explorable space and distribute robots to unknown areas.

The method, as shown in Alg.~\ref{alg:linear_convex_hull}, is a heuristic approach that starts by figuring out the convex hull of the observed map. Observing wall cells in the exploration map on the convex hull is used to forecast potential unobserved areas of the area. The probabilistic Hough line transformation, which uses a maximum likelihood estimation of a line via sparsely connected points, is used to find these unobserved sections of the region. It identifies unseen connections between observed wall segments. The map's unseen areas are then expanded along the identified wall lines.

Each junction of these wall-line projections adds a projected corner (represented by a single cell, $c_{i,j} \in I_o$) to the exploration map. Limiting the distance of predicted corners from the closest observations is necessary to prevent nearly parallel lines from predicting corner locations that are unreasonably far from the observed space.
The observed areas of the map and the corner points are then combined to form a new convex hull.
This new convex hull provides an initial estimation of the boundaries of the exploration space. The cells inside the hull are initially anticipated to be free, while the cells along the hull are predicted to be occupied in the exploration map.

It is possible that the anticipated boundary, which is convex, overestimates the limits of the explorable space. However, this upbeat strategy boosts the payout in hazy places, increasing the motivation for their exploration. Although extending lines suggests that most buildings have a rectangular shape, this is consistent with many areas humans have created. While external borders are considered straight, the projected boundaries are not constrained to any orientation or polygons class, allowing the prediction to adapt to non-rectangular situations.

In the case of a non-linearize convex hull, projecting robots to specific unexplored regions would be non-optimal. To linearize a non-convex hull, we first need to find a convex hull that approximates the non-convex function. This can be done using techniques such as convex relaxation or piecewise linearization. Once we have a convex hull approximation, we can use the following linearization equation:
\begin{equation}
f(C) \geq \sum_{j=1}^{n} a_j \nabla f_j(C) + b ,
\end{equation}
where $f(C)$ is the non-convex function we want to optimize, $ \nabla f_j(C)$ are the convex functions that approximate $f(C)$, $a_j$ are non-negative coefficients, and $b$ is a constant that ensures the approximation is valid.

Upkeep on the projected map is done before moving on to the whole exploration. To account for unseen depth in barriers, occupied cells, $O_o \cup I_o$, are inflated into $I_f$; for example, when an obstacle is only visible from one side, it has no observable depth. User-specified inflation depth is determined by prior operating environment knowledge. Then, the map's unreachable regions inside the expected perimeter are labeled as inferred obstacles.

\subsection{Time Complexity}
\label{sec:complexity}
We examine the temporal complexity for each observation using Algs.~\ref{alg:SEAL} and~\ref{alg:linear_convex_hull}. The map resolution $\alpha_M$, the map size $n$, the maximum size of a convex hull $C$, the number of beams per scan $n_z$, the number of measurement $\alpha_z$, and the one of continuous map occupancy $\alpha_m$ for a map of size $m$ is used to define the complexity. We assume that the squared grid's size is fixed, the query time is constant, and the data structure containing the OG map is identical to the GP data structure.

The proposed SEAL has two parallel processes: Gaussian process fusion for exploration belief estimation and Graph optimization for belief estimation of relative localization. Rao-Blackwellization will use both beliefs to update overall localization and map beliefs. The resulting time complexity of two parallel processes are $O(n^2)$ and $O(\alpha^{-2}_Mn_z\alpha^{-1}_z\alpha^{-1}_m)$. It is worth noting that the time complexities of standard Log-Odds for generating OG mapping from GP are $O(n_zlogn)$. Note that $n_z<n$ normally because of the limited sensing range and narrow beam angle. The time complexity of the proposed SEAL is, at worst, quadratic about the cone size and map resolution for belief update and GP computation, respectively. Importantly, the time cost of map belief update is quite similar to Log-Odds approach in a large scene because the independence of map size $n$ and the actual cone size is much smaller in cluttered environments. The time efficiency of SEAL calculation also depends on the linearization of convex hull optimization, which is $O(nlogn)$. Hence, the overall upper bound on the time complexity per robot for the proposed SEAL will be $O(n\times m)$, where m is the number of immediate (connected) neighbor robots.

This time complexity is significantly better than other methods, such as SLAM + RRT \cite{zhang2020rapidly} and DRL \cite{hu2020voronoi}, which have complexities  $O\left(n\times \log (n)\right)$ and $O(n\times k)$, respectively, where $k$ is the number of temporal states, and $m << k$.

\begin{figure*}[t]
\centering
 \includegraphics[width=0.33\linewidth]{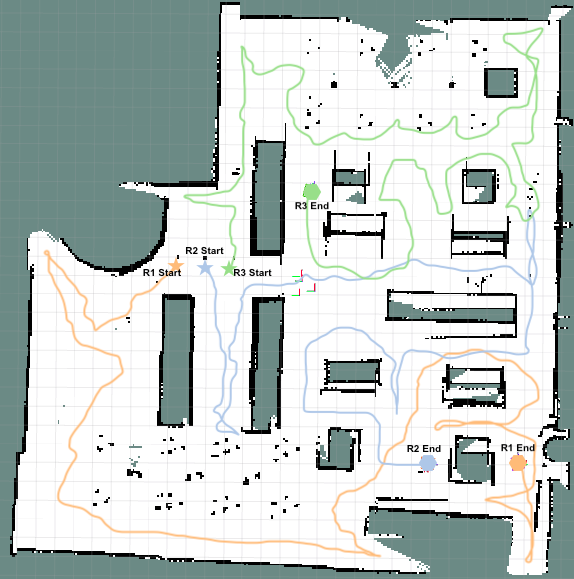}
 \includegraphics[width=0.3\linewidth]{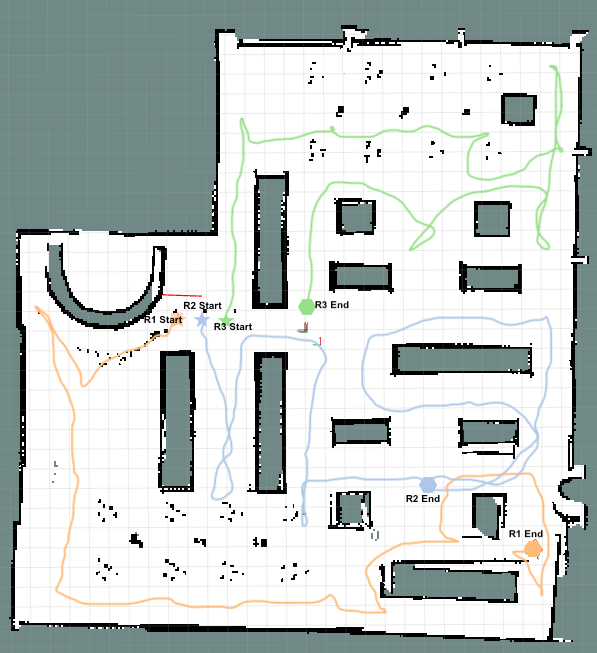}
 \includegraphics[width=0.32\linewidth]{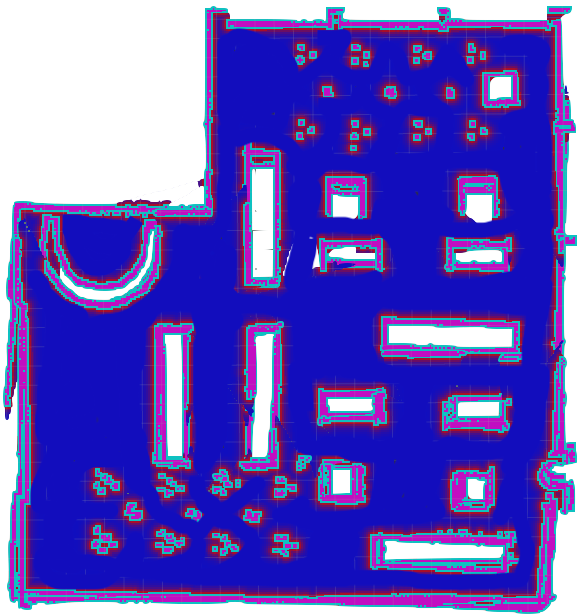}
 
 \vspace{2mm}
 \includegraphics[width=0.3\linewidth]{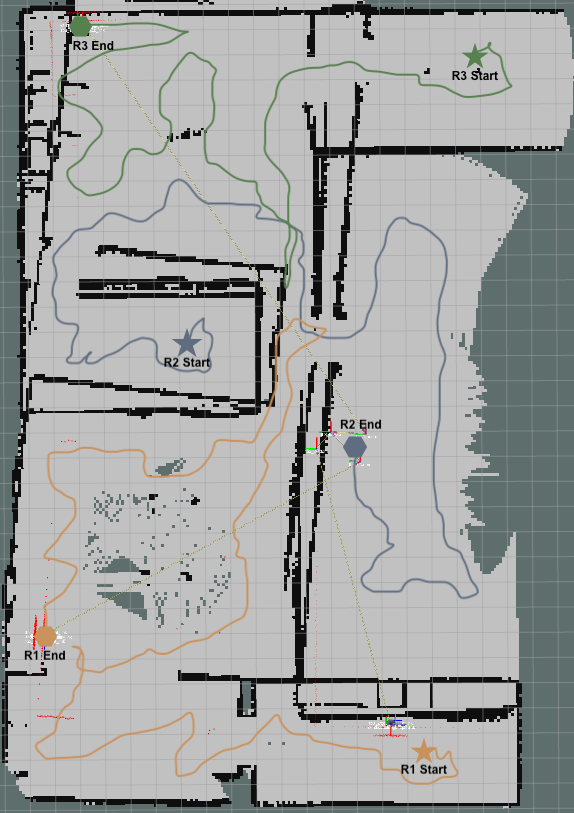}
 \includegraphics[width=0.3\linewidth]{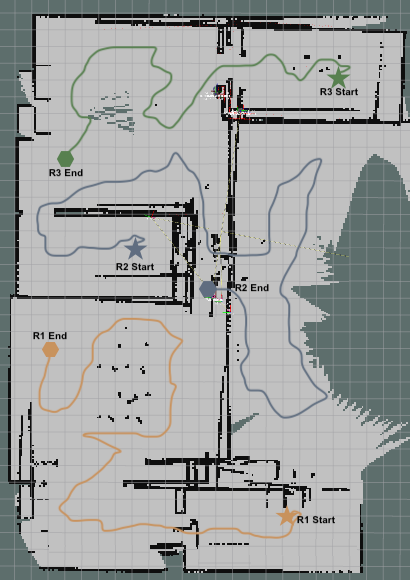}
 \includegraphics[width=0.32\linewidth]{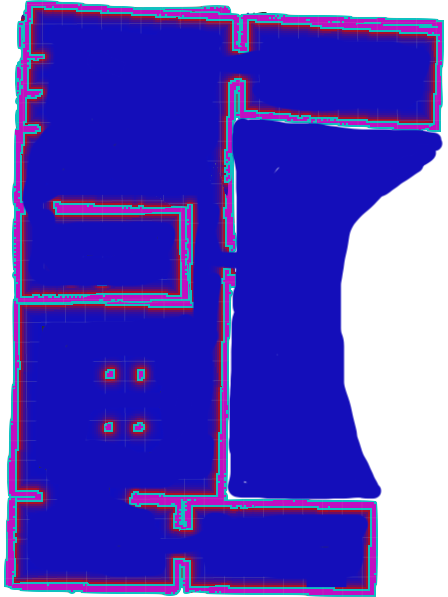}
\caption{Exploration grid maps along trajectories: The top pictures show the Bookstore world, and the bottom pictures show the House world. \textbf{(Top Left)} Explored map using RRT \cite{zhang2020rapidly}, \textbf{(Top Center)} Explored map using proposed SEAL, \textbf{(Top Right)} Explored maps' confidence (belief map) of SEAL. \textbf{(Bottom Left)} Explored map using DRL \cite{hu2020voronoi}, \textbf{(Bottom Center)} Explored map using proposed SEAL, \textbf{(Bottom Right)} Explored maps' confidence (belief map) of SEAL. \textit{Confidence maps are provided only by SEAL.}}
 \label{fig:maps}
\end{figure*}

\begin{figure}[t]
\centering
 \includegraphics[width=0.49\linewidth]{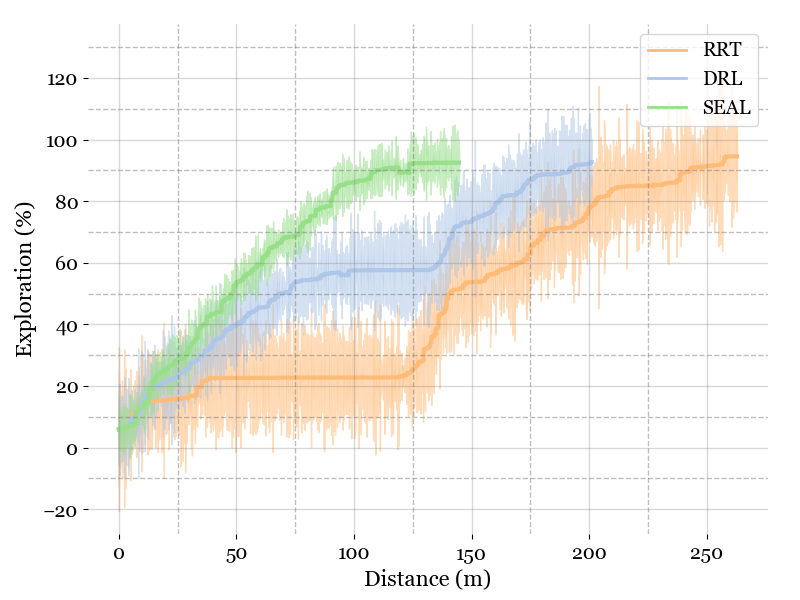}
  \includegraphics[width=0.49\linewidth]{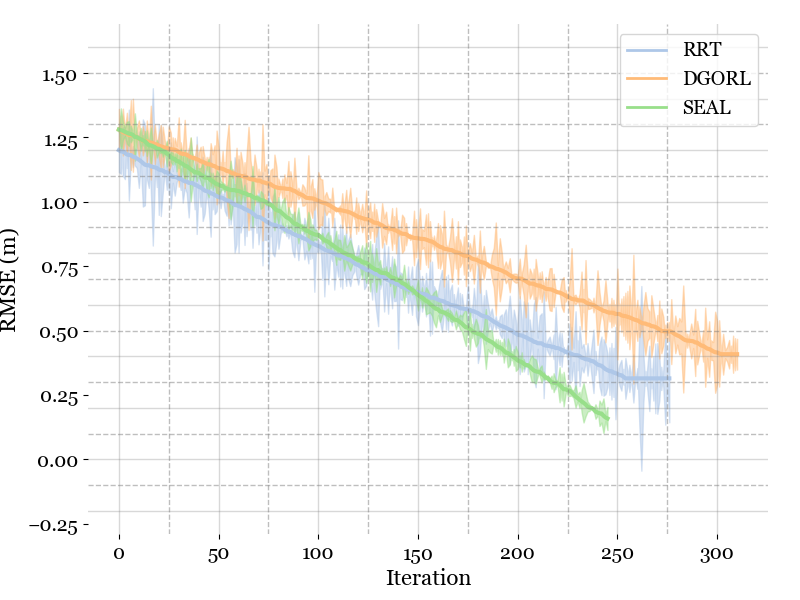}
\caption{Exploration progress \textbf{(Left)} and localization error \textbf{(Right)} of the proposed SEAL, RRT \cite{zhang2020rapidly}, and DRL \cite{hu2020voronoi} based exploration and mapping.}
 \label{fig:explore_distance}
 \vspace{-4mm}
\end{figure}

\begin{figure}[t]
\centering
 \includegraphics[width=0.36\linewidth]{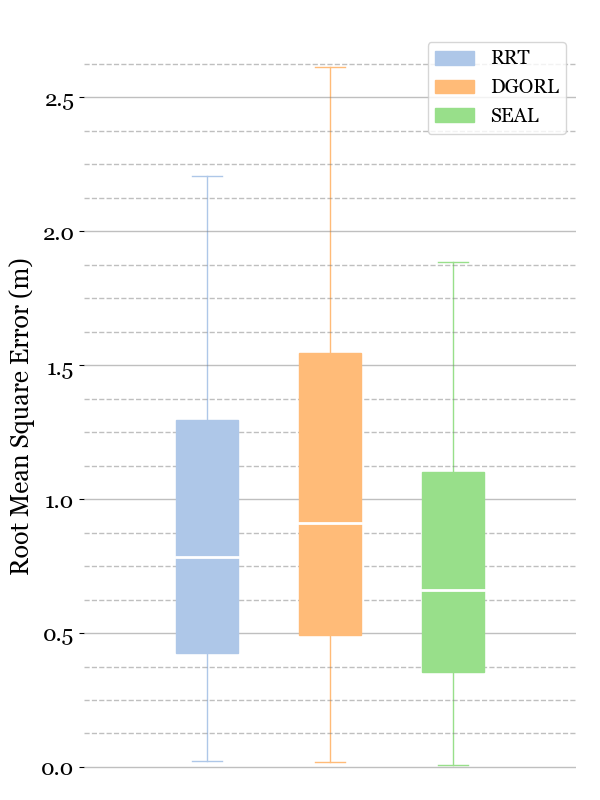}
 \includegraphics[width=0.62\linewidth]{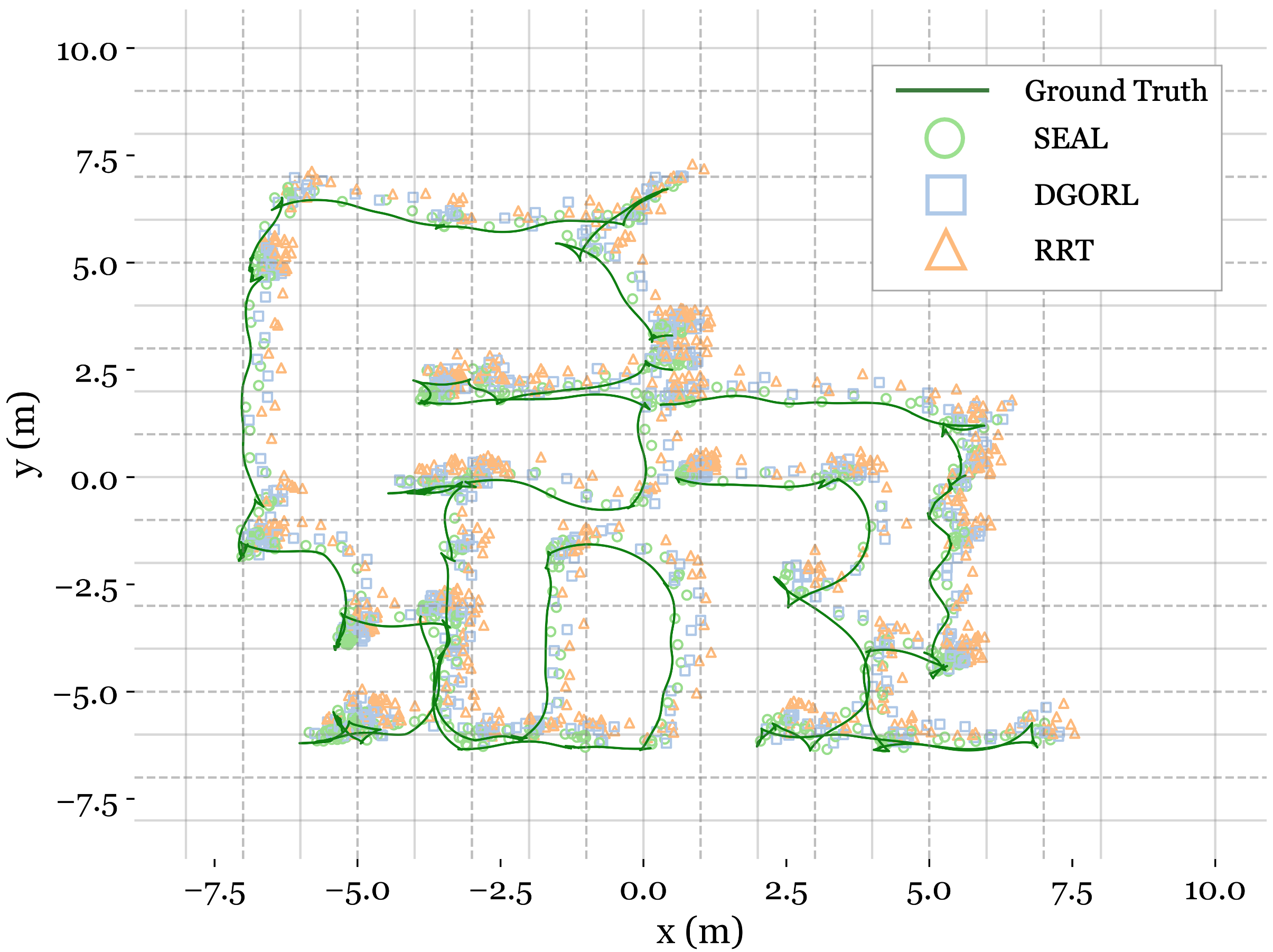}
   \vspace{-2mm}
 \hspace{-2mm}
\caption{\textbf{(Left)} Localization error comparison with RRT \cite{zhang2020rapidly} and DGORL \cite{latif2022dgorl}, \textbf{(Right)} Predicted trajectory of the three robots for different approaches.}
 \label{fig:loc_trajectory}
\end{figure}

\begin{table*}[t]
\begin{center}
\caption{Performance results of different approaches. $^*$ indicates the best performer. L - Localization. E - Exploration.}
\label{exp_results}
\begin{tabular}{|c|c|c|c|c|}
\hline
\multirow{3}{*}{\textbf{Parameters}} &\textbf{SLAM + RRT } & \textbf{DRL } & \textbf{DGORL } & \textbf{SEAL} \\
~ & \cite{zhang2020rapidly} & \cite{hu2020voronoi}  & \cite{latif2022dgorl} &  (ours) \\
~ &L + E & E Only  & L Only & L + E \\
\hline
\multicolumn{5}{|c|}{\textbf{Exploration Metrics}} \\
\hline
\textbf{Mapping Time (s)} & $302\pm22$ & $273\pm33$ & N/A & $\textbf{247}^*\pm19$\\
\hline
 \textbf{Total distance (m)} & $267\pm14$ & $198\pm21$ & N/A & $\textbf{139}^*\pm11$ \\ 
\hline
\textbf{Explored Area (\%)} & $92\pm2$ & $94\pm3$ & N/A & $\textbf{97}^*\pm2$ \\
\hline
\multicolumn{5}{|c|}{\textbf{Localization and Mapping Metrics}} \\ \hline
\textbf{Map SSIM} & $0.69\pm0.15$ & $0.72\pm0.12$ & N/A & $\textbf{0.88}^*\pm0.09$ \\
\hline
\textbf{ATE (m)} & $0.5\pm0.09$ & N/A & $0.09\pm0.04$ & $\textbf{0.04}^*\pm0.02$\\
\hline
\textbf{ALE (m)} & $1.3\pm0.14$ & N/A & $0.81\pm0.5$ & $\textbf{0.63}^*\pm0.2$ \\ 
\hline
\multicolumn{5}{|c|}{\textbf{Computing, Communication, and Efficiency Metrics}} \\ \hline
\textbf{CPU (\%)} & $109\pm20$  & $88\pm17$  & N/A & $\textbf{82}^*\pm11$ \\ 
\hline
\textbf{RAM (MB)} & $1014\pm22$  &  $1321\pm59$ & N/A & $\textbf{891}^*\pm27$  \\ 
\hline
\textbf{Communication (MB)} & $2.4\pm0.07$  &  $2.7\pm0.09$ & N/A & $\textbf{1.2}^* \pm 0.04$  \\ 
\hline
\end{tabular}
\end{center}
\end{table*}

\section{Experimental Validation}
\label{sec:experiments}
We evaluated the SEAL approach in two situations in ROS Gazebo-modified versions of the AWS bookstore and house worlds.  A laser sensor with a $180^\circ$ field of vision, a 5 m range, and 1500 beams per scan are installed on the robot's wheeled platform. The robot's maximum speeds in both directions are 0.2 $\frac{m}{s}$ and 0.8 $\frac{rad}{s}$, respectively. Each robot in the experimentation performs SEAL and generates a map based on updated belief for the GP grid using Rao-Blackwellization (discussed in Section~\ref{sec:rao-blackwell}). Each robot computes linearized convex hull optimization for navigation, subsequently moving to the closest boundary based on the temporal convex hull.

\subsection{Evaluation Metrics}
\textbf{Exploration:} The exploration approach and the method put forward by RRT \cite{zhang2020rapidly} and DRL \cite{hu2020voronoi} are compared with that of SEAL in the bookstore and house worlds, respectively, since they use the same environments so that we can obtain a fair comparison. Below are the used metrics for evaluation:
\begin{enumerate}
    \item \textbf{Mapping Time}: The amount of time spent in the mapping (a measure of how effective is the multi-robot exploration); 
    \item \textbf{Total Distance Traveled:} This term refers to the cumulative path length of all robot's trajectory, which is necessary for a multi-robot system to explore the entire map. The entire trajectory length gives an idea of the robot's energy usage while subtly describing its investigation's effectiveness;
    \item \textbf{Exploration Percentage:} The percentage of the generated map with time elapsed;
    \item \textbf{Map SSIM:} Structural similarity index measure of generated maps compared with ground truth map;
    \item \textbf{CPU Utilization:} The maximum percentage consumption of the processor of a robot throughout the robot's trajectory;
    \item \textbf{Memory Consumption (RAM):} The maximum occupied memory by the robot throughout the trajectory;
    \item \textbf{Communication payload (Data Size):} The maximum size of shared data by a robot.
\end{enumerate}
\textbf{Localization:} The localization approach used RRT-Exploration \cite{zhang2020rapidly} (which used GMapping for SLAM and Map merging to merge maps and find relative localization between robots) and DGORL \cite{latif2022dgorl} (which only does relative localization) are compared in our experiments. We use the below metrics for evaluation:
\begin{enumerate}
    \item \textbf{ATE (m)}: The absolute trajectory error computed for the whole navigated trajectory by each approach, which measured the deviation from the ground truth;
    \item \textbf{ALE (m):} The absolute localization error for the predicted location of each approach.
\end{enumerate}

\subsection{Exploration Results}
Comprehensive listing of all results are provided in Table~\ref{exp_results}. 
To reduce measurement noise, we examined the exploration performance of each strategy after several trials. There are several simulations run in a three-robot system.
Fig.~\ref{fig:maps} displays the mapping results of each approach made by the three robots in the simulated area. Fig.~\ref{fig:maps} also delineates the belief maps generated by SEAL, which are not provided by typical SLAM (or map merging) algorithms or other exploration approaches. 

Note that the results consider the typical mapping time, the distance traveled, and the effectiveness of the mapping. Comparison with the original map is used to gauge mapping effectiveness and reported percentages are adjusted using gazebo world dimensions. Fig.~\ref{fig:explore_distance} delineates the reduced traveled distance and achieves a high exploration percentage. It can be seen that the proposed SEAL traveled approximately half the distance and achieved a comparable exploration percentage. Furthermore, The exploration capabilities of RRT, DRL, and SEAL are thoroughly compared in Table~\ref{exp_results}. These results suggest that SEAL is a more effective method for exploration activities, producing superior outcomes.

\subsection{Localization Results}
We also have analyzed the localization performance of SOTA RRT and our previous approach DGORL for localization and compared results with the proposed SEAL. Fig.~\ref{fig:loc_trajectory} has shown the trajectory by each approach and localization error in terms of RMSE. Results show that our SEAL algorithm outperformed the RRT Exploration and DGORL, as it incorporates the map belief for location correction and achieved 25\% higher localization accuracy than DGORL and 10\% higher than RRT. 
To analyze the localization performance over iterations, Fig.~\ref{fig:explore_distance} provides evidence for a gradual reduction in RMSE over robotic progression towards the maximum exploration. Additionally, Table~\ref{exp_results} provided exact data showing the significant improvement of the absolute trajectory and localization errors by SEAL compared to the state-of-the-art approaches. Specifically, the proposed SEAL has 25\% higher localization accuracy than RRT and DGORL.

\section{Conclusion}
This paper proposes a novel simultaneous exploration and localization (SEAL) approach for distributed multi-robot systems by integrating Gaussian Processes and graph optimization techniques.
SEAL outperformed in exploration and localization performances compared to other state-of-the-art exploration and localization methods. Additionally, compared to RRT and DRL, SEAL decreased the time complexity per robot by around 35\% and 29\%, respectively, and achieved comparable mapping efficiency with improved localization accuracy than RRT and DGORL. SEAL offers fast and precise exploration without depending on boundary information for localization, making it especially beneficial for various robotic applications; as it does not rely on external sensors, it is appropriate for dynamic settings with few features or no prior knowledge.

\bibliography{ref}

\begin{thebibliography}{10}
\providecommand{\url}[1]{#1}
\csname url@samestyle\endcsname
\providecommand{\newblock}{\relax}
\providecommand{\bibinfo}[2]{#2}
\providecommand{\BIBentrySTDinterwordspacing}{\spaceskip=0pt\relax}
\providecommand{\BIBentryALTinterwordstretchfactor}{4}
\providecommand{\BIBentryALTinterwordspacing}{\spaceskip=\fontdimen2\font plus
\BIBentryALTinterwordstretchfactor\fontdimen3\font minus
  \fontdimen4\font\relax}
\providecommand{\BIBforeignlanguage}[2]{{%
\expandafter\ifx\csname l@#1\endcsname\relax
\typeout{** WARNING: IEEEtran.bst: No hyphenation pattern has been}%
\typeout{** loaded for the language `#1'. Using the pattern for}%
\typeout{** the default language instead.}%
\else
\language=\csname l@#1\endcsname
\fi
#2}}
\providecommand{\BIBdecl}{\relax}
\BIBdecl

\bibitem{xu2021crmi}
Y.~Xu, R.~Zheng, M.~Liu, and S.~Zhang, ``Crmi: Confidence-rich mutual
  information for information-theoretic mapping,'' \emph{IEEE Robotics and
  Automation Letters}, vol.~6, no.~4, pp. 6434--6441, 2021.

\bibitem{latif2022dgorl}
E.~Latif and R.~Parasuraman, ``Dgorl: Distributed graph optimization based
  relative localization of multi-robot systems,'' \emph{arXiv preprint
  arXiv:2210.01662}, 2022.

\bibitem{botteghi2020entropy}
M.~Botteghi, M.~Khaled, B.~Sirmacek, and M.~Poel, ``Entropy-based exploration
  for mobile robot navigation: a learning-based approach,'' in \emph{Planning
  and robotics workshop, PlanRob}, 2020.

\bibitem{jadidi2015mutual}
M.~G. Jadidi, J.~V. Miro, and G.~Dissanayake, ``Mutual information-based
  exploration on continuous occupancy maps,'' in \emph{2015 IEEE/RSJ
  International Conference on Intelligent Robots and Systems (IROS)}.\hskip 1em
  plus 0.5em minus 0.4em\relax IEEE, 2015, pp. 6086--6092.

\bibitem{bai2016information}
S.~Bai, J.~Wang, F.~Chen, and B.~Englot, ``Information-theoretic exploration
  with bayesian optimization,'' in \emph{2016 IEEE/RSJ International Conference
  on Intelligent Robots and Systems (IROS)}.\hskip 1em plus 0.5em minus
  0.4em\relax IEEE, 2016, pp. 1816--1822.

\bibitem{valencia2018active}
R.~Valencia, J.~Andrade-Cetto, R.~Valencia, and J.~Andrade-Cetto, ``Active pose
  slam,'' \emph{Mapping, Planning and Exploration with Pose SLAM}, pp. 89--108,
  2018.

\bibitem{schmuck2019ccm}
P.~Schmuck and M.~Chli, ``Ccm-slam: Robust and efficient centralized
  collaborative monocular simultaneous localization and mapping for robotic
  teams,'' \emph{Journal of Field Robotics}, vol.~36, no.~4, pp. 763--781,
  2019.

\bibitem{latif2022multi}
E.~Latif and R.~Parasuraman, ``Multi-robot synergistic localization in dynamic
  environments,'' in \emph{ISR Europe 2022; 54th International Symposium on
  Robotics}.\hskip 1em plus 0.5em minus 0.4em\relax VDE, 2022, pp. 1--8.

\bibitem{zhang2020rapidly}
L.~Zhang, Z.~Lin, J.~Wang, and B.~He, ``Rapidly-exploring random trees
  multi-robot map exploration under optimization framework,'' \emph{Robotics
  and Autonomous Systems}, vol. 131, p. 103565, 2020.

\bibitem{hu2020voronoi}
J.~Hu, H.~Niu, J.~Carrasco, B.~Lennox, and F.~Arvin, ``Voronoi-based
  multi-robot autonomous exploration in unknown environments via deep
  reinforcement learning,'' \emph{IEEE Transactions on Vehicular Technology},
  vol.~69, no.~12, pp. 14\,413--14\,423, 2020.

\bibitem{pham2009convex}
V.~Pham, C.~Laird, and M.~El-Halwagi, ``Convex hull discretization approach to
  the global optimization of pooling problems,'' \emph{Industrial \&
  Engineering Chemistry Research}, vol.~48, no.~4, pp. 1973--1979, 2009.

\bibitem{rao2008rao}
C.~R. Rao, ``Rao-blackwell theorem,'' \emph{Scholarpedia}, vol.~3, no.~8, p.
  7039, 2008.

\bibitem{hornung2013octomap}
A.~Hornung, K.~M. Wurm, M.~Bennewitz, C.~Stachniss, and W.~Burgard, ``Octomap:
  An efficient probabilistic 3d mapping framework based on octrees,''
  \emph{Autonomous robots}, vol.~34, pp. 189--206, 2013.

\bibitem{zhang2020fsmi}
Z.~Zhang, T.~Henderson, S.~Karaman, and V.~Sze, ``Fsmi: Fast computation of
  shannon mutual information for information-theoretic mapping,'' \emph{The
  International Journal of Robotics Research}, vol.~39, no.~9, pp. 1155--1177,
  2020.

\bibitem{shi2020adaptive}
Y.~Shi, N.~Wang, J.~Zheng, Y.~Zhang, S.~Yi, W.~Luo, and K.~Sycara, ``Adaptive
  informative sampling with environment partitioning for heterogeneous
  multi-robot systems,'' in \emph{2020 IEEE/RSJ International Conference on
  Intelligent Robots and Systems (IROS)}.\hskip 1em plus 0.5em minus
  0.4em\relax IEEE, 2020, pp. 11\,718--11\,723.

\bibitem{placed2021fast}
J.~A. Placed and J.~A. Castellanos, ``Fast autonomous robotic exploration using
  the underlying graph structure,'' in \emph{2021 IEEE/RSJ International
  Conference on Intelligent Robots and Systems (IROS)}.\hskip 1em plus 0.5em
  minus 0.4em\relax IEEE, 2021, pp. 6672--6679.

\bibitem{swiler2020survey}
L.~P. Swiler, M.~Gulian, A.~L. Frankel, C.~Safta, and J.~D. Jakeman, ``A survey
  of constrained gaussian process regression: Approaches and implementation
  challenges,'' \emph{Journal of Machine Learning for Modeling and Computing},
  vol.~1, no.~2, 2020.

\bibitem{wang2022fast}
Z.~Wang, L.~Chen, H.~Chen, H.~Chen, and X.~Jiang, ``Fast and safe exploration
  via adaptive semantic perception in outdoor environments,'' in \emph{2022
  IEEE/RSJ International Conference on Intelligent Robots and Systems
  (IROS)}.\hskip 1em plus 0.5em minus 0.4em\relax IEEE, 2022, pp. 9445--9451.

\bibitem{gan2020bayesian}
L.~Gan, R.~Zhang, J.~W. Grizzle, R.~M. Eustice, and M.~Ghaffari, ``Bayesian
  spatial kernel smoothing for scalable dense semantic mapping,'' \emph{IEEE
  Robotics and Automation Letters}, vol.~5, no.~2, pp. 790--797, 2020.

\bibitem{jadidi2014exploration}
M.~G. Jadidi, J.~V. Mir{\'o}, R.~Valencia, and J.~Andrade-Cetto, ``Exploration
  on continuous gaussian process frontier maps,'' in \emph{2014 IEEE
  International Conference on Robotics and Automation (ICRA)}.\hskip 1em plus
  0.5em minus 0.4em\relax IEEE, 2014, pp. 6077--6082.

\bibitem{ghaffari2018gaussian}
M.~Ghaffari~Jadidi, J.~Valls~Miro, and G.~Dissanayake, ``Gaussian processes
  autonomous mapping and exploration for range-sensing mobile robots,''
  \emph{Autonomous Robots}, vol.~42, pp. 273--290, 2018.

\bibitem{chen2020autonomous}
F.~Chen, J.~D. Martin, Y.~Huang, J.~Wang, and B.~Englot, ``Autonomous
  exploration under uncertainty via deep reinforcement learning on graphs,'' in
  \emph{2020 IEEE/RSJ International Conference on Intelligent Robots and
  Systems (IROS)}.\hskip 1em plus 0.5em minus 0.4em\relax IEEE, 2020, pp.
  6140--6147.

\bibitem{chen2021zero}
F.~Chen, P.~Szenher, Y.~Huang, J.~Wang, T.~Shan, S.~Bai, and B.~Englot,
  ``Zero-shot reinforcement learning on graphs for autonomous exploration under
  uncertainty,'' in \emph{2021 IEEE International Conference on Robotics and
  Automation (ICRA)}.\hskip 1em plus 0.5em minus 0.4em\relax IEEE, 2021, pp.
  5193--5199.

\bibitem{robert2021rao}
C.~P. Robert and G.~O. Roberts, ``Rao-blackwellization in the mcmc era,''
  \emph{arXiv preprint arXiv:2101.01011}, 2021.

\bibitem{luo2018adaptive}
W.~Luo and K.~Sycara, ``Adaptive sampling and online learning in multi-robot
  sensor coverage with mixture of gaussian processes,'' in \emph{2018 IEEE
  International Conference on Robotics and Automation (ICRA)}.\hskip 1em plus
  0.5em minus 0.4em\relax IEEE, 2018, pp. 6359--6364.

\bibitem{parasuraman2014multi}
R.~Parasuraman, T.~Fabry, L.~Molinari, K.~Kershaw, M.~Di~Castro, A.~Masi, and
  M.~Ferre, ``A multi-sensor rss spatial sensing-based robust stochastic
  optimization algorithm for enhanced wireless tethering,'' \emph{Sensors},
  vol.~14, no.~12, pp. 23\,970--24\,003, 2014.

\end{thebibliography}
\bibliographystyle{IEEEtran}

\end{document}